\pdfoutput=1

\documentclass[11pt]{article}
\usepackage[table]{xcolor}
\usepackage[final]{acl}
\usepackage{array}
\usepackage{cite}
\usepackage{rotating}
\usepackage{bigstrut}
\usepackage{subcaption}
\usepackage{booktabs,rotating,caption, amsmath, amssymb}
\usepackage{amsfonts,amssymb}
\usepackage{times}  
\usepackage{booktabs}
\usepackage{xspace}
\usepackage{amsmath}
\usepackage{enumitem}
\usepackage{tcolorbox}
\usepackage{tabularx}

\usepackage{multirow}
\usepackage[T1]{fontenc}
\usepackage[utf8]{inputenc}  
\usepackage{pdfpages} 
\usepackage{microtype}  
\usepackage{inconsolata}  
\usepackage{hyperref}
\usepackage{cleveref}
\usepackage{latexsym}  
\usepackage{tabularx}
\usepackage{graphicx}
\usepackage{pifont}
\usepackage{adjustbox}
\newcommand{\CheckmarkBold}{\checkmark}
\newcommand{\XSolidBrush}{\ding{55}}
\newcommand{\eg}[0]{\emph{e.g.},~}
\newcommand{\method}[1]{\textsc{#1}}
\newcommand{\ie}[0]{\emph{i.e.},~}
\newcommand{\llamaa}{\textsc{LLaMA2-7B-Chat}\xspace}
\newcommand{\llamab}{\textsc{LLaMA2-13B-Chat}\xspace}
\newcommand{\gpta}{\textsc{GPT-3.5-Turbo-0613}\xspace}
\newcommand{\vicuna}{\textsc{Vicuna-7B-v1.5}\xspace}
\newcommand{\qwen}{\textsc{Qwen2-7B}\xspace}
\makeatletter
\newcommand{\ssymbol}[1]{^{\@fnsymbol{#1}}}
\makeatother
\newcommand{\mistral}{\textsc{Mistral-7B}\xspace}
\definecolor{lightgray}{gray}{0.95}

\title{Can Graph Descriptive Order Affect Solving  Graph Problems with LLMs?}

\author{
    Yuyao Ge$^{1}$\thanks{~~The authors are also affiliated with: Key Laboratory of Network Data Science and Technology, Institute of Computing Technology, Chinese Academy of Sciences; State Key Laboratory of AI Safety; University of Chinese Academy of Sciences.}\quad Shenghua Liu$^{1}$\footnotemark[1]\thanks{~~Corresponding author.}\quad Baolong Bi$^{1}$\footnotemark[1]\quad Yiwei Wang$^{2}$\quad \\
    \textbf{Lingrui Mei}$^{1}$\footnotemark[1] \textbf{Wenjie Feng}$^3$\quad \textbf{Lizhe Chen}$^4$\quad \textbf{Xueqi Cheng}$^{1}$\footnotemark[1] \\
    $^1$Institute of Computing Technology, Chinese Academy of Sciences \\    
    $^2$University of California, Merced \\
    $^3$SAIDS, University of Science and Technology of China \\
    $^4$Tsinghua University \\
    {\{geyuyao24z, liushenghua\}@ict.ac.cn} 
}
\makeatletter
\renewcommand\@makefntext[1]{%
    \noindent\makebox[-1pt][l]{\@makefnmark\hspace{0em}}#1}
\makeatother

\begin{document}
	
	\maketitle
	\begin{abstract}
            Large language models (LLMs) have achieved significant success in reasoning tasks, including mathematical reasoning and logical deduction. 
            Among these reasoning tasks, graph problems stand out due to their complexity and unique structural characteristics, attracting considerable attention from researchers.
            Previous studies have explored LLMs' graph reasoning abilities through various techniques, such as different encoding methods for graph structures and the use of carefully designed prompts. 
            However, a critical factor has been mostly overlooked: the prompt sequential order in which graph descriptions are presented to the models.
            In this study, we present the first comprehensive analysis of how the order of graph descriptions impacts LLM performance.
            Specifically, we comprehensively evaluate four graph description orders across six graph problems using six mainstream LLMs. 
            The results reveal that: (1) ordered graph descriptions significantly improve LLMs' comprehension of graph structures; (2) the robustness of LLMs to graph description order varies across different tasks; and (3) the impact of graph order on performance is closely related to the inherent characteristics of tasks. 
            This study provides a critical advancement in the application of LLMs for solving graph-related problems, paving the way for future research to optimize model performance through strategic graph description ordering. 
	\end{abstract}


        
            \section{Introduction}
    Large language models (LLMs) have made remarkable progress, showing unprecedented capabilities in NLP~\citep{vaswani2017attention, devlin2018bert, brown2020language, ouyang2022training}. Leveraging advancements in NLP, LLMs excel in reasoning tasks, which has drawn considerable interest from researchers. As a type of complex reasoning problem, graph problems have also attracted substantial attention. For instance, \citet{wang2023can} represented graphs in natural language and validated the effectiveness of prompts for graph reasoning tasks. \citet{fatemi2023talk} map pure graphs in a real-world scenario to understand how LLMs’ learned representations are leveraged in graph tasks. More recently, \citet{skianis2024graphreasoninglargelanguage} explore the use of pseudo-code instructions to enhance LLMs' ability to solve graph problems.

    \begin{figure}[t]
        \centering
        \includegraphics[width=\linewidth]{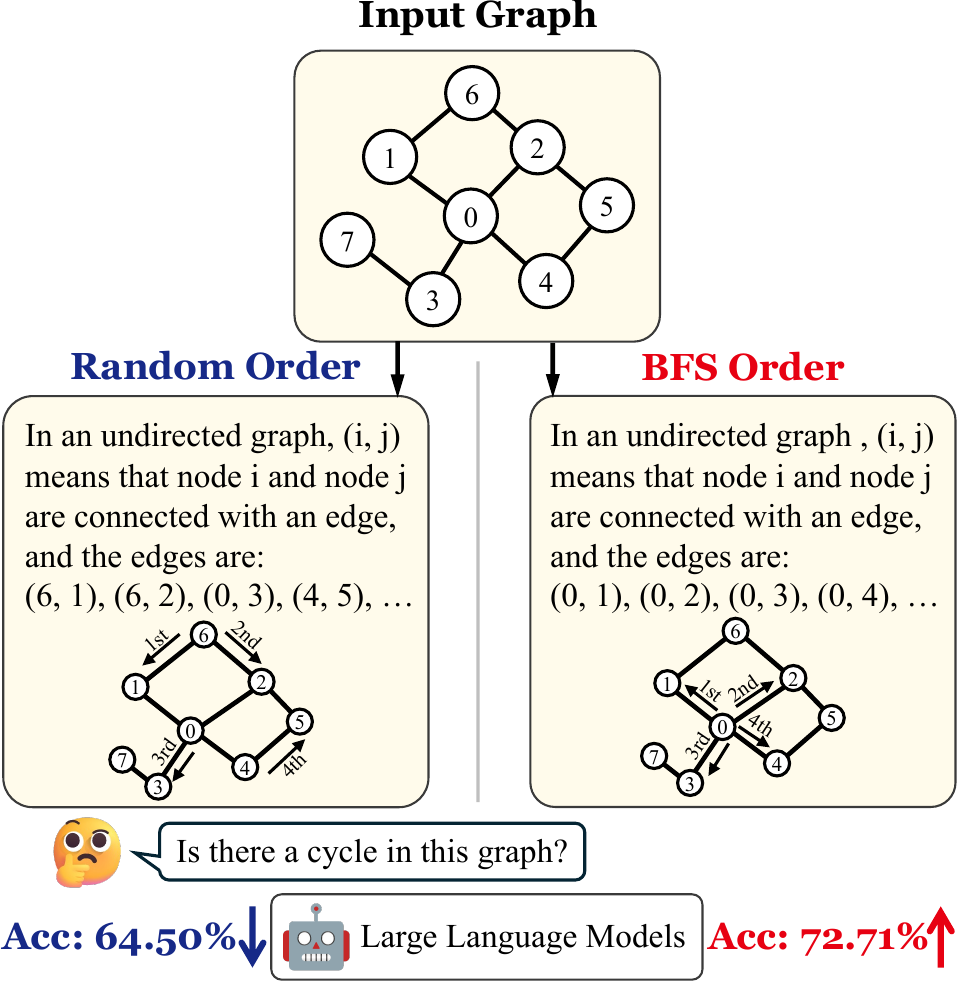} 
        \caption{The order in which graphs are described significantly affects LLMs’ ability to understand and solve graph problems. For instance, in the cycle detection task, graphs described in BFS order achieved an average accuracy improvement of 12.73\% over those described in random order.} 
        \label{pdf:7}
    \end{figure}
	
    \par 
    
    \begin{figure*}[tbp]\vspace{-0.8cm}
        \centering
        \includegraphics[width=1\linewidth]{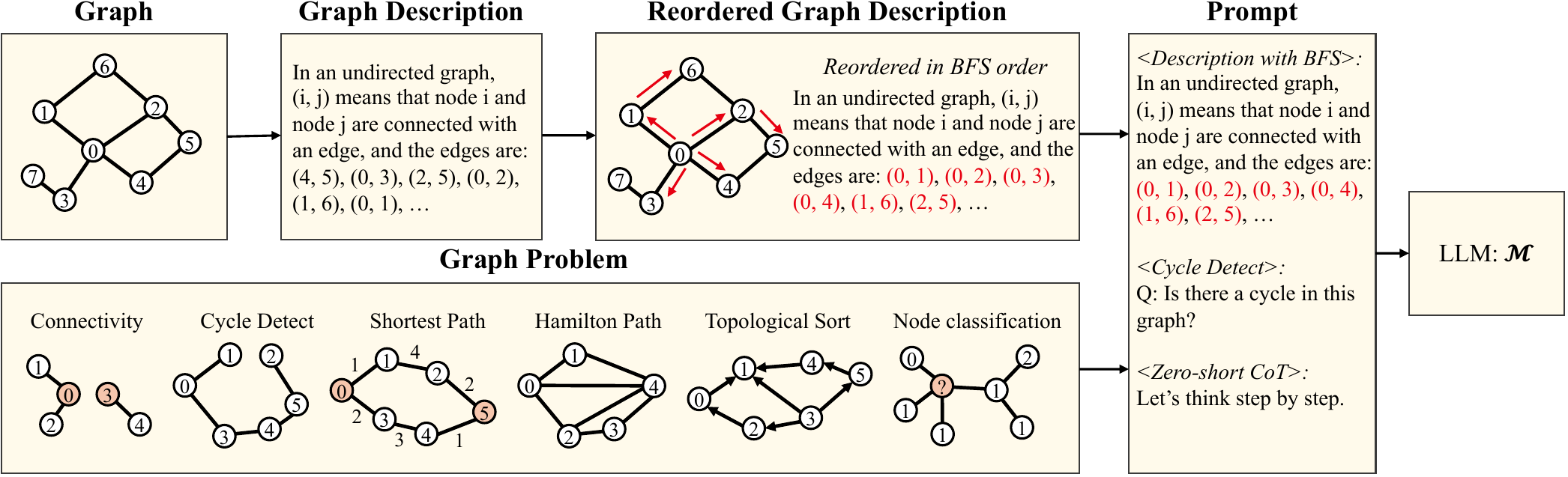} 
        \caption{Overview of our framework for solving graph problems with LLMs. \textit{In node classification task, node labels no longer represent identifiers; instead, they indicate the categories the nodes belong to.}} 
        \label{pdf:1}
    \end{figure*}

    Despite the significant contributions of previous researchers, one key issue remains overlooked: \textbf{the order of graph descriptions may affect LLMs' performance in solving graph problems}. They typically employed randomly arranged graph descriptions, overlooking the critical role that description order may play. While graphs have no fixed textual representation, the order in which their components are expressed may affect the model's reasoning process. Different orders of descriptions may emphasize specific paths or parts, providing LLMs with different perspectives on the same graph structure. 

    In this work, we explore the impact of the order of graph description in solving graph problems with LLMs. For comprehensive study, we design four graph description orders and categorized them into graph traversal-based orders, including BFS and DFS, and probability distribution-based orders, including PageRank and Personalized PageRank. These orders were carefully chosen to provide LLMs different perspectives on graph understanding: BFS provides a hierarchical traversal, DFS offers a deep traversal, PageRank delivers a global probability distribution of node importance, and Personalized PageRank focuses on a localized probability distribution. Given that certain orders may influence specific graph reasoning tasks, we design six graph tasks spanning varying levels of complexity, and we conducted experiments on six mainstream LLMs.
    
    Our main contributions are as follows:
    
    \begin{itemize}
    [noitemsep,topsep=0pt,parsep=0pt,partopsep=0pt,leftmargin=1em]
        \item We are the first to demonstrate that the order of graph descriptions significantly affects the graph reasoning performance of LLMs.
        \item Through extensive experimentation, we analyzed the differential impact of description orders on LLMs' performance across diverse graph reasoning tasks.
        \item We introduced the GraphDO (\textbf{Graph} \textbf{D}escription with \textbf{O}rder), a novel dataset consisting of a set of graphs, corresponding prompts, and predefined description orders, which aims to advance the community’s understanding of how graph description impacts reasoning in LLMs.
    \end{itemize}

        \section{Preliminary}
	
    \subsection{Prompt Engineering for Graph}	
    Prompt engineering involves strategically designing task-specific instructions, referred to as prompts, to guide model output without altering parameters~\citep{sahoo2024systematic, collobert2008unified, mikolov2013efficient, sutskever2014sequence}. 
    
    In this work, we consider a graph \( \mathcal{G} = (\mathcal{V}, \mathcal{E}) \), where \( \mathcal{V} \) denotes the set of nodes and \( \mathcal{E} \) the set of edges. For encoding graph to text, we define the graph encoding function \( g(\mathcal{G}, o) \), which maps a graph \( \mathcal{G} \) and a description order \( o \in \mathcal{O} \) to a description in natural language. Additionally, let \( q(T) \) be a function that generates a question \( Q \) based on a graph task \( T \), such that \( q: T \mapsto Q \), where \( Q \) has a standard answer \( Y \). Graph prompting engineering can be formally expressed as an optimization problem focused on determining the optimal order \( o \) that maximizes the LLM $\mathcal{M}$'s scoring function $\mathcal{S}$, and is formulated as:
    \begin{equation}
        \label{eq:2}
        \max_{o \in \mathcal{O}} \; \mathbb{E}_{\mathcal{G}, T, Y \in D} \; \mathcal{S}(\mathcal{M}(p, g(\mathcal{G}, o), q(T)), Y)
    \end{equation}
    where $p \in \mathcal{P}$ represents the prompt style, $D$ is the dataset consisting of triples $(\mathcal{G}, T, Y)$.
    
    To ensure the completeness of our experiments, we set five prompt styles to offer varying levels of heuristic reasoning to LLMs, with more details in Appendix \ref{sec:prompt}. 
    
    \subsection{Graph Problems} \label{sec:task}
    We designed six graph reasoning tasks, covering various levels of reasoning complexity and categories.
    
\paragraph{T1 Connectivity} In an undirected graph, LLMs need to determine whether a path exists between two arbitrary nodes \( u, v \in \mathcal{V} \), which is a binary classification problem.
    
\paragraph{T2 Cycle} In an undirected graph, LLMs need to determine if a non-empty path exists where the starting and ending nodes are the same, which is a binary classification problem.
    
\paragraph{T3 Hamilton Path} A Hamilton path visits each node in \( \mathcal{V} \) exactly once. The LLMs need to determine whether such a path exists in an undirected graph \( \mathcal{G} \) and, if such a path exists, answer the path.

\paragraph{T4 Shortest Path} In an undirected graph, LLMs need to answer the complete shortest path between two nodes \( u, v \in \mathcal{V} \).
    
\paragraph{T5 Topological Sort} In a directed graph \( \mathcal{G} \), LLMs need to generate a linear ordering of the nodes such that for every directed edge \( (u, v) \in \mathcal{E} \), node \( u \) precedes node \( v \) in the ordering. This task requires finding any valid topological sort of the graph, and multiple correct solutions may exist.
    
\paragraph{T6 Node Classification} In an undirected graph composed of nodes with labels, LLMs need to predict the label of a certain node which is labeled as '?' based on the labels of its neighbouring nodes. There is only one node labeled as '?' in each graph.

The T1-T5 tasks focus on pure graph structures to evaluate LLMs' understanding of graphs. Specifically, T1 and T2 assess local reasoning, while T3 to T5 examine global graph understanding. T6 shifts the focus to graph attribute learning, making it more relevant for real-world applications.

        \crefname{section}{§}{§§}
\Crefname{section}{§}{§§}

\section{Graph Description Generation}

\subsection{Graph Encoder}
Graphs can be described in text through multiple encoding methods.  \citet{fatemi2023talk} compared several approaches for converting graph data into text sequences. Given that our research focuses on the order of graph descriptions, we adopted the adjacency format, which uses edge lists to represent graphs and can be applied to both pure and attributed graphs, making it ideal for our study.

To encode an edge list \( \mathcal{L}_o \) into a graph description in adjacency format, we use a template function \( \mathcal{T}(\cdot) \). The process is formalized as:

\begin{equation}
    \label{eq:g}
    g(\mathcal{G}, o) = \mathcal{T}( \mathcal{G}, \mathcal{L}_o), o \in \mathcal{O}
\end{equation}

For pure unweighted graphs (\ie cycle detection, connectivity detection, shortest path, Hamilton path), the definition of \( \mathcal{T}(\cdot) \) function is as follows:

\begin{tcolorbox}[fonttitle = \small\bfseries, title=Prompt Template for Unweighted Graphs,colframe=gray!2!black,colback=gray!2!white,boxrule=1pt,boxsep=0pt,left=5pt,right=5pt,fontupper=\footnotesize, halign title = flush center]
In an undirected/directed graph, ({\color[RGB]{71,162,26}{\textit{i}}}, {\color[RGB]{229,91,11}{\textit{j}}}) means that node {\color[RGB]{71,162,26}{\textit{i}}} and node {\color[RGB]{229,91,11}{\textit{j}}} are connected with an edge, and the edges are: [(0, 1), (1, 3), (3, 5), \dots].
\end{tcolorbox}

For pure weighted graphs (\ie shortest path), we define the \( \mathcal{T}(\cdot) \) as:

\begin{tcolorbox}[fonttitle = \small\bfseries, title=Prompt Template for Weighted Graphs,colframe=gray!2!black,colback=gray!2!white,boxrule=1pt,boxsep=0pt,left=5pt,right=5pt,fontupper=\footnotesize, halign title = flush center]
In an undirected/directed graph, ({\color[RGB]{71,162,26}{\textit{i}}}, {\color[RGB]{229,91,11}{\textit{j}}}, {\color[RGB]{2, 76, 170}{\textit{w}}}) means that node {\color[RGB]{71,162,26}{\textit{i}}} and node {\color[RGB]{229,91,11}{\textit{j}}} are connected by an edge with weight {\color[RGB]{2, 76, 170}{\textit{w}}}, and the edges are: [(1, 3, 2), (0, 3, 1), (0, 1, 4), \dots].
\end{tcolorbox}

For graphs composed of nodes with labels (\ie node classification), inspired by \citet{das2023modality}, we define the \( \mathcal{T}(\cdot) \) as:

\begin{tcolorbox}[fonttitle = \small\bfseries, title=Prompt Template for Node Classification Task,colframe=gray!2!black,colback=gray!2!white,boxrule=1pt,boxsep=0pt,left=5pt,right=5pt,fontupper=\footnotesize, halign title = flush center]
Adjacency list: [(1758, 2217), (2217, 2645), \ldots]

Node to label mapping: node 1758: label 3 | node 2217: label 2 | node 2645: label ? | \ldots

\end{tcolorbox}

\subsection{Graph Description Ordering} \label{sec: Graph Description Ordering}

Beyond the random order, we designed four additional graph description orders for our main experiment. Furthermore, two more description orders were employed in the deeper exploration, which will be discussed in the corresponding section.

\paragraph{Random Order} In random order, the edges \( \mathcal{E} \) of the graph are shuffled randomly. In previous works, researchers have commonly employed random graph description orders (\eg~\citep{wang2023can, fatemi2023talk, das2023modality}).

\paragraph{Breadth-First Search (BFS) Order} Starting from a random root node \( v_0 \in \mathcal{V} \), BFS generates a sequence of edges by exploring the graph level by level. At each level, for each node \( v \), the edges \( (v, u) \), where \( u \in \mathcal{N}(v) \), are added to the sequence before moving to the next level. 

\paragraph{Depth-First Search (DFS) Order} Starting from a node \( v_0 \in \mathcal{V} \), DFS follows a recursive strategy, generating a sequence of edges by traversing as deeply as possible before backtracking. 

\paragraph{PageRank (PR) Order} In PR order, nodes \( v \in \mathcal{V} \) are sorted in descending order by their PageRank scores \( PR(v) \), where \( PR(v) = \alpha \sum_{u \in \mathcal{N}^{-1}(v)} \frac{PR(u)}{|\mathcal{N}(u)|} + (1 - \alpha) \), with \( \alpha = 0.85 \) as the damping factor and \( \mathcal{N}^{-1}(v) \) as the nodes linking to \( v \). For each node, starting with the highest-ranked, edges to its neighbors \( u \in \mathcal{N}(v) \) are added to the edge list $\mathcal{L}_{PR}$. If an edge \( (v, u) \) or \( (u, v) \) is already in $\mathcal{L}_{PR}$, it is skipped.

\paragraph{Personalized PageRank (PPR) Order} PPR introduces a personalization vector mechanism that prioritizes proximity to specific target nodes. The ranking is computed as \( PR_S(v) = \alpha \sum_{u \in \mathcal{N}^{-1}(v)} \frac{PR_S(u)}{|\mathcal{N}(u)|} + (1 - \alpha) \cdot e_v \), where the parameter \( e_v \) is task-specific and its definition can be found in the Appendix \ref{pv}. The subsequent computations follow the same process as in PageRank.

\textbf{Note.} For DFS and BFS order, when traversing the graph \( \mathcal{G} \), it is not guaranteed that all edges \( e \in \mathcal{E} \) will be included in the edge list \( \mathcal{L} \). To avoid this, we perform a traversal on the dual graph \( \mathcal{G}^* \) of \( \mathcal{G} \) to ensure that the resulting edge list includes all the edges in \( \mathcal{G} \). For a disconnected graph, the root node will be reselected randomly until the graph is fully described. This method does not alter the topology of \( \mathcal{G} \); it merely serves as a means of obtaining \( \mathcal{L} \).

\begin{table*}[!t]
\small
\renewcommand{\arraystretch}{0.9}
\centering
\setlength{\tabcolsep}{2pt}
\setlength{\extrarowheight}{4pt} 
\resizebox{\textwidth}{!}{%
\begin{tabular}{c|c|p{2cm}<{\centering}p{2cm}<{\centering}p{2cm}<{\centering}p{2cm}<{\centering}p{2cm}<{\centering} p{2cm}<{\centering}}
\toprule[1.5pt]
\textbf{Task} & \textbf{Order} & \textbf{Zero-shot} & \textbf{Zero-shot CoT} & \textbf{Few-shot} & \textbf{CoT} & \textbf{CoT-BAG} & \textbf{Avg.} \\ 
\midrule

\multirow{5}{*}{\begin{sideways}\footnotesize \method{Conn.}\end{sideways}}
& \cellcolor{gray!25} Random & \cellcolor{gray!25} 73.93$_{\left(\text{\small -}\right)}$ & \cellcolor{gray!25} 70.71$_{\left(\text{\small -}\right)}$ & \cellcolor{gray!25} 81.07$_{\left(\text{\small -}\right)}$ & \cellcolor{gray!25} 83.93$_{\left(\text{\small -}\right)}$ & \cellcolor{gray!25} 82.14$_{\left(\text{\small -}\right)}$ & \cellcolor{gray!25} 78.36$_{\left(\text{\small -}\right)}$ \\
& BFS & 82.14$_{\left(\uparrow 11.11\right)}$ & 87.50$_{\left(\uparrow 23.74\right)}$ & 89.29$_{\left(\uparrow 10.14\right)}$ & 92.50$_{\left(\uparrow 10.21\right)}$ & 95.71$_{\left(\uparrow 16.52\right)}$ & 89.43$_{\left(\uparrow 14.13\right)}$ \\
& DFS & 79.29$_{\left(\uparrow 7.25\right)}$ & 82.14$_{\left(\uparrow 16.16\right)}$ & 87.14$_{\left(\uparrow 7.49\right)}$ & 88.21$_{\left(\uparrow 5.10\right)}$ & 89.29$_{\left(\uparrow 8.70\right)}$ & 85.21$_{\left(\uparrow 8.75\right)}$ \\
& PR & 77.86$_{\left(\uparrow 5.32\right)}$ & 83.57$_{\left(\uparrow 18.19\right)}$ & 85.71$_{\left(\uparrow 5.72\right)}$ & 84.29$_{\left(\uparrow 0.43\right)}$ & 87.50$_{\left(\uparrow 6.53\right)}$ & 83.79$_{\left(\uparrow 6.93\right)}$ \\
& PPR & 76.79$_{\left(\uparrow 3.87\right)}$ & 81.07$_{\left(\uparrow 14.65\right)}$ & 83.93$_{\left(\uparrow 3.53\right)}$ & 84.64$_{\left(\uparrow 0.85\right)}$ & 86.07$_{\left(\uparrow 4.78\right)}$ & 82.50$_{\left(\uparrow 5.29\right)}$ \\ 
\midrule

\multirow{5}{*}{\begin{sideways}\footnotesize \method{Cycle}\end{sideways}} 
& \cellcolor{gray!25} Random & \cellcolor{gray!25} 51.79$_{\left(\text{\small -}\right)}$ & \cellcolor{gray!25} 53.57$_{\left(\text{\small -}\right)}$ & \cellcolor{gray!25} 65.36$_{\left(\text{\small -}\right)}$ & \cellcolor{gray!25} 75.71$_{\left(\text{\small -}\right)}$ & \cellcolor{gray!25} 76.07$_{\left(\text{\small -}\right)}$ & \cellcolor{gray!25} 64.50$_{\left(\text{\small -}\right)}$ \\
& BFS & 55.71$_{\left(\uparrow 7.57\right)}$ & 56.07$_{\left(\uparrow 4.67\right)}$ & 79.29$_{\left(\uparrow 21.31\right)}$ & 86.07$_{\left(\uparrow 13.68\right)}$ & 86.43$_{\left(\uparrow 13.62\right)}$ & 72.71$_{\left(\uparrow 12.73\right)}$ \\
& DFS & 52.14$_{\left(\uparrow 0.68\right)}$ & 53.93$_{\left(\uparrow 0.67\right)}$ & 73.21$_{\left(\uparrow 12.01\right)}$ & 79.29$_{\left(\uparrow 4.73\right)}$ & 81.07$_{\left(\uparrow 6.57\right)}$ & 67.93$_{\left(\uparrow 5.31\right)}$ \\ 
& PR & 55.36$_{\left(\uparrow 6.89\right)}$ & 56.43$_{\left(\uparrow 5.33\right)}$ & 70.36$_{\left(\uparrow 7.65\right)}$ & 80.36$_{\left(\uparrow 6.14\right)}$ & 83.21$_{\left(\uparrow 9.39\right)}$ & 69.14$_{\left(\uparrow 7.20\right)}$ \\
& PPR & 54.29$_{\left(\uparrow 4.83\right)}$ & 55.00$_{\left(\uparrow 2.67\right)}$ & 70.00$_{\left(\uparrow 7.10\right)}$ & 79.29$_{\left(\uparrow 4.73\right)}$ & 80.00$_{\left(\uparrow 5.17\right)}$ & 67.72$_{\left(\uparrow 4.99\right)}$ \\ 
\midrule

\multirow{5}{*}{\begin{sideways}\footnotesize \method{HamPath}\end{sideways}} 
& \cellcolor{gray!25} Random & \cellcolor{gray!25} 10.71$_{\left(\text{\small -}\right)}$ & \cellcolor{gray!25} 15.36$_{\left(\text{\small -}\right)}$ & \cellcolor{gray!25} 40.00$_{\left(\text{\small -}\right)}$ & \cellcolor{gray!25} 46.07$_{\left(\text{\small -}\right)}$ & \cellcolor{gray!25} 45.36$_{\left(\text{\small -}\right)}$ & \cellcolor{gray!25} 31.50$_{\left(\text{\small -}\right)}$ \\
& BFS & 20.00$_{\left(\uparrow 86.74\right)}$ & 20.71$_{\left(\uparrow 34.83\right)}$ & 57.86$_{\left(\uparrow 44.65\right)}$ & 58.57$_{\left(\uparrow 27.13\right)}$ & 57.14$_{\left(\uparrow 25.97\right)}$ & 42.86$_{\left(\uparrow 36.05\right)}$ \\
& DFS & 33.93$_{\left(\uparrow 216.81\right)}$ & 37.50$_{\left(\uparrow 144.14\right)}$ & 67.50$_{\left(\uparrow 68.75\right)}$ & 63.93$_{\left(\uparrow 38.77\right)}$ & 59.29$_{\left(\uparrow 30.71\right)}$ & 52.43$_{\left(\uparrow 66.44\right)}$ \\ 
& PR & 15.00$_{\left(\uparrow 40.06\right)}$ & 19.29$_{\left(\uparrow 25.59\right)}$ & 48.93$_{\left(\uparrow 22.32\right)}$ & 55.00$_{\left(\uparrow 19.38\right)}$ & 50.00$_{\left(\uparrow 10.23\right)}$ & 37.64$_{\left(\uparrow 19.50\right)}$ \\
& PPR & 16.43$_{\left(\uparrow 53.41\right)}$ & 18.93$_{\left(\uparrow 23.24\right)}$ & 50.00$_{\left(\uparrow 25.00\right)}$ & 53.93$_{\left(\uparrow 17.06\right)}$ & 50.36$_{\left(\uparrow 11.02\right)}$ & 37.93$_{\left(\uparrow 20.41\right)}$ \\ 
\midrule

\multirow{5}{*}{\begin{sideways}\footnotesize \method{TopoSort}\end{sideways}}  
& \cellcolor{gray!25} Random & \cellcolor{gray!25} 28.93$_{\left(\text{\small -}\right)}$ & \cellcolor{gray!25} 31.07$_{\left(\text{\small -}\right)}$ & \cellcolor{gray!25} 58.21$_{\left(\text{\small -}\right)}$ & \cellcolor{gray!25} 56.07$_{\left(\text{\small -}\right)}$ & \cellcolor{gray!25} 60.36$_{\left(\text{\small -}\right)}$ & \cellcolor{gray!25} 46.93$_{\left(\text{\small -}\right)}$ \\
& BFS & 43.21$_{\left(\uparrow 49.36\right)}$ & 40.36$_{\left(\uparrow 29.90\right)}$ & 67.14$_{\left(\uparrow 15.34\right)}$ & 61.43$_{\left(\uparrow 9.56\right)}$ & 65.00$_{\left(\uparrow 7.69\right)}$ & 55.43$_{\left(\uparrow 18.11\right)}$ \\
& DFS & 42.14$_{\left(\uparrow 45.66\right)}$ & 48.93$_{\left(\uparrow 57.48\right)}$ & 77.86$_{\left(\uparrow 33.76\right)}$ & 74.29$_{\left(\uparrow 32.50\right)}$ & 72.86$_{\left(\uparrow 20.71\right)}$ & 63.21$_{\left(\uparrow 34.71\right)}$ \\ 
& PR & 35.36$_{\left(\uparrow 22.23\right)}$ & 35.71$_{\left(\uparrow 14.93\right)}$ & 71.07$_{\left(\uparrow 22.09\right)}$ & 58.21$_{\left(\uparrow 3.82\right)}$ & 65.36$_{\left(\uparrow 8.28\right)}$ & 53.14$_{\left(\uparrow 13.24\right)}$ \\
& PPR & 37.14$_{\left(\uparrow 28.38\right)}$ & 39.64$_{\left(\uparrow 27.58\right)}$ & 72.50$_{\left(\uparrow 24.55\right)}$ & 58.93$_{\left(\uparrow 5.10\right)}$ & 66.43$_{\left(\uparrow 10.06\right)}$ & 54.93$_{\left(\uparrow 17.05\right)}$ \\ 
\midrule

\multirow{5}{*}{\begin{sideways}\footnotesize \method{SPath}\end{sideways}} 
& \cellcolor{gray!25} Random & \cellcolor{gray!25} 20.00$_{\left(\text{\small -}\right)}$ & \cellcolor{gray!25} 25.00$_{\left(\text{\small -}\right)}$ & \cellcolor{gray!25} 26.07$_{\left(\text{\small -}\right)}$ & \cellcolor{gray!25} 38.93$_{\left(\text{\small -}\right)}$ & \cellcolor{gray!25} 40.71$_{\left(\text{\small -}\right)}$ & \cellcolor{gray!25} 30.14$_{\left(\text{\small -}\right)}$ \\
& BFS & 35.36$_{\left(\uparrow 76.80\right)}$ & 42.50$_{\left(\uparrow 70.00\right)}$ & 45.36$_{\left(\uparrow 73.99\right)}$ & 67.50$_{\left(\uparrow 73.39\right)}$ & 65.71$_{\left(\uparrow 61.41\right)}$ & 51.29$_{\left(\uparrow 70.15\right)}$ \\
& DFS & 32.14$_{\left(\uparrow 60.70\right)}$ & 34.29$_{\left(\uparrow 37.16\right)}$ & 45.00$_{\left(\uparrow 72.61\right)}$ & 58.57$_{\left(\uparrow 50.45\right)}$ & 57.14$_{\left(\uparrow 40.36\right)}$ & 45.43$_{\left(\uparrow 50.71\right)}$ \\
& PR & 30.36$_{\left(\uparrow 51.80\right)}$ & 43.93$_{\left(\uparrow 75.72\right)}$ & 38.93$_{\left(\uparrow 49.33\right)}$ & 43.93$_{\left(\uparrow 12.84\right)}$ & 48.93$_{\left(\uparrow 20.19\right)}$ & 41.21$_{\left(\uparrow 36.74\right)}$ \\
& PPR & 32.50$_{\left(\uparrow 62.50\right)}$ & 44.64$_{\left(\uparrow 78.56\right)}$ & 42.14$_{\left(\uparrow 61.64\right)}$ & 45.36$_{\left(\uparrow 16.52\right)}$ & 49.64$_{\left(\uparrow 21.94\right)}$ & 42.86$_{\left(\uparrow 42.18\right)}$ \\ 
\bottomrule[1.5pt]
\end{tabular}
}
\caption{Results of the performance of various orders on different graph tasks. ${\left(\uparrow\right)}$ indicates the improvement compared to the baseline under the same setting.}
\label{table:2}
\end{table*}

        \section{Experiments}

\subsection{Experimental Settings}

\paragraph{Datasets}

Our experiments are conducted on the GraphDO dataset, introduced in this paper, which includes six graph tasks. GraphDO consists of 8,500 cases, with each case containing a graph description, a question, and an answer. Each graph description is generated in a specific order. For traditional graph tasks (e.g., cycle detection, connectivity detection, shortest path, Hamilton path, topological sort), we employ the Erdős-Rényi (ER) graph generation method. We apply a filtering process to the generated graphs to ensure that each case has a valid and well-defined solution. For the graph learning task (e.g., node classification), we conduct experiments on attributed graphs using three widely recognized datasets: CORA \citep{mccallum2000automating}, Citeseer \citep{giles1998citeseer}, and Pubmed \citep{sen2008collective}. Since the sizes of these real-world citation graphs exceed the input limits of LLMs, we employ graph sampling methods, including ego-graph (Ego) and forest fire sampling (FF). Since the node classification task requires fewer reasoning steps than traditional graph tasks, we set the default prompt style to zero-shot. Additional details about the GraphDO dataset can be found in Appendix \ref{sec:odgraph}.

\paragraph{Models and Settings}
We use the \gpta as the default model \citep{brown2020language}. To ensure the generality of our conclusions, we also conducte experiments on other models, including \llamaa, \llamab \citep{touvron2023llama2openfoundation}, \qwen \citep{qwen2}, \mistral \citep{jiang2023mistral7b}, and \vicuna \citep{zheng2023judging}. The decoding temperature is set to zero.

\paragraph{Metric}
Performance is measured by accuracy, defined as:
    \begin{equation}
        \label{eq:4}
		\text{Acc} = \frac{\#\textit{correct answers}}{\#\textit{total questions}}
    \end{equation}
where \# represents the number of instances.

More judgment details are in Appendix \ref{sec:parser}.

\paragraph{Baseline}
We use the random order graph description as a baseline to facilitate comparisons with ordered descriptions.

\subsection{Main Result}

\paragraph{(Q1) Does the order of graph description impact the LLM’s performance in solving graph problems?} 

\begin{figure}[h] 
    \centering
    \includegraphics[width=\linewidth]{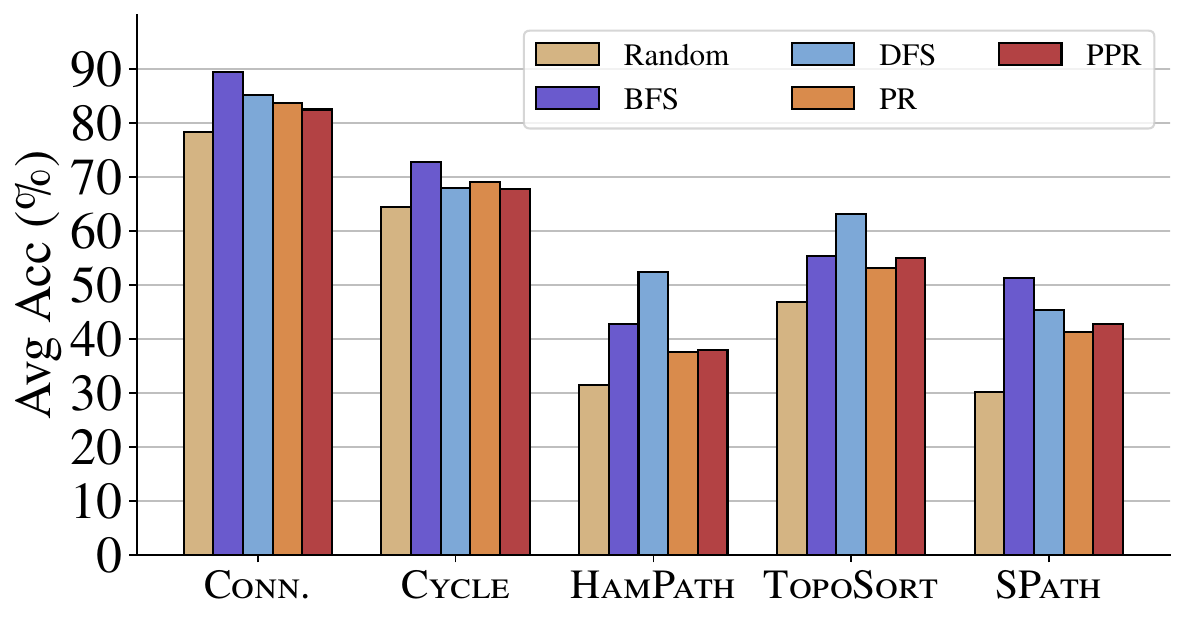}
    \caption{The LLM’s average accuracy in solving various tasks across different orders.} 
    \label{pdf:23}
\end{figure}

As presented in Table~\ref{table:2}, ordered graph descriptions consistently outperform the random baseline across all traditional graph tasks and prompt configurations. For instance, in the connectivity task, the BFS order achieves an average accuracy of 89.43\%, significantly higher than the random order’s 78.36\%. Similarly, in the cycle detection task, the BFS order reaches an accuracy of 72.71\%, compared to the random order’s 64.50\%. Figure~\ref{pdf:23} further illustrates that the random order consistently yields the lowest accuracy across tasks.

As presented in Table~\ref{table:3}, ordered descriptions also consistently outperform the baseline in node classification task. Specifically, in the CORA dataset with ego-graph sampling, PR order achieves 75.33\% accuracy, compared to 70.00\% for the random order. Similarly, in the Pubmed dataset, PR order reaches 82.67\%, which significantly surpasses the random order's 72.00\%.

\begin{figure}[h] \hspace{0cm}
    \centering
    \includegraphics[width=\linewidth]{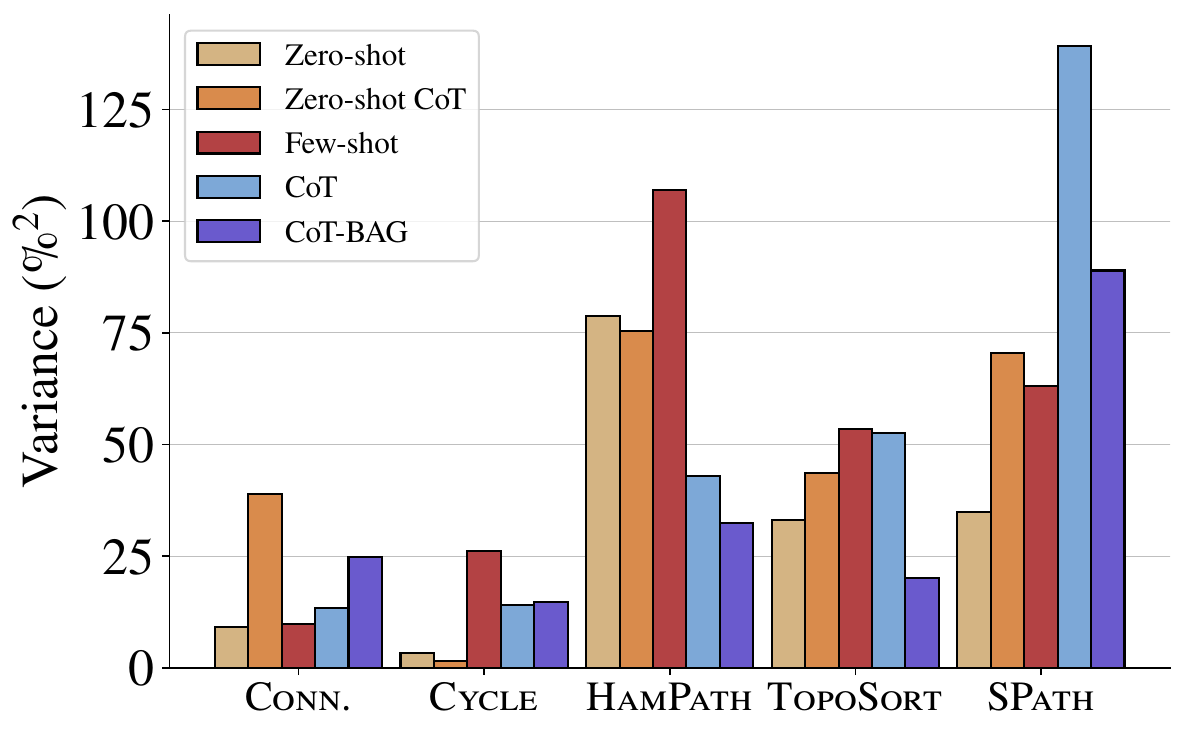}
    \caption{Variance of LLM accuracy across different graph tasks with varying description orders. The variance for each task is computed as \( \sigma^2 = \frac{1}{|\mathcal{O}|} \sum_{o \in \mathcal{O}} \left( \mathcal{S}_o - \mu \right)^2 \), where \( \mathcal{S}_o \) is the accuracy for order \( o \), \( \mu \) is the mean accuracy across all orders.} 
    \label{pdf:10}
\end{figure}

However, the improvements in node classification task is generally less pronounced than those observed in traditional graph tasks. For example, in the Pubmed dataset with ego-graph sampling, the PR order improves accuracy by 14.82\%, which is the largest improvment in node classification tasks. This suggests that, while ordered descriptions are indeed advantageous, their relatively smaller impact on node classification tasks may be attributed to inherent differences in the complexity and reasoning patterns between these tasks.

We hypothesize that LLMs' improved performance with ordered descriptions stems from limitations in positional encoding and attention mechanisms—what we call attention bias. Positional encodings, which are meant to provide sequence information in transformer models, may not effectively capture the structural complexity of graph data when the input sequence is unordered. Guided by positional encodings, attention mechanisms can give undue priority to certain sections of the input based on their order in the sequence, leading to over-dependence on the graph description.

\begin{table*}[!t] \vspace{-0.3cm}
\centering
\small
\renewcommand{\arraystretch}{0.9} 
\setlength{\tabcolsep}{2pt}
\setlength{\extrarowheight}{4pt} 
\resizebox{\textwidth}{!}{%
\begin{tabular}{c|c|p{2.1cm}<{\centering} p{2.1cm}<{\centering} | p{2.1cm}<{\centering} p{2.1cm}<{\centering} | p{2.1cm}<{\centering} p{2.1cm}<{\centering}} 

\toprule[1.5pt]
\multirow{2}{*}{\textbf{Sampling}} & \multirow{2}{*}{\textbf{Order}} & \multicolumn{2}{c|}{\textbf{CORA}} & \multicolumn{2}{c|}{\textbf{Citeseer}} & \multicolumn{2}{c}{\textbf{Pubmed}} \\  

 &  & \textbf{Acc.} & \(\Delta\) & \textbf{Acc.} & \(\Delta\) & \textbf{Acc.} & \(\Delta\) \\
\midrule

\multirow{5}{*}{Ego}
& \cellcolor{gray!25} Random & \cellcolor{gray!25}70.00 & \cellcolor{gray!25}- & \cellcolor{gray!25}67.33 & \cellcolor{gray!25}- & \cellcolor{gray!25}72.00 & \cellcolor{gray!25}- \\ 
& BFS & 72.00 & ${\uparrow 2.86 }$ & 68.67 & ${\uparrow 1.99 }$ & 74.00 & ${\uparrow 2.78 }$ \\ 
& DFS & 71.33 & ${\uparrow 1.90 }$ & 68.66 & ${\uparrow 1.98 }$ & 77.33 & ${\uparrow 7.40 }$ \\
& PR & 75.33 & ${\uparrow 7.61 }$ & 71.33 & ${\uparrow 5.94 }$ & 82.67 & ${\uparrow 14.82 }$ \\
& PPR & 73.33 & ${\uparrow 4.76 }$ & 69.33 & ${\uparrow 2.97 }$ & 77.33 & ${\uparrow 7.40 }$ \\ 
\midrule

\multirow{5}{*}{Forest Fire}
& \cellcolor{gray!25} Random & \cellcolor{gray!25}79.33 & \cellcolor{gray!25}- & \cellcolor{gray!25}68.67 & \cellcolor{gray!25}- & \cellcolor{gray!25}69.99 & \cellcolor{gray!25}- \\
& BFS & 82.67 & ${\uparrow 4.21 }$ & 71.33 & ${\uparrow 3.87 }$ & 74.00 & ${\uparrow 5.73 }$ \\
& DFS & 81.33 & ${\uparrow 2.52 }$ & 70.00 & ${\uparrow 1.94 }$ & 76.00 & ${\uparrow 8.59 }$ \\ 
& PR & 83.33 & ${\uparrow 5.04 }$ & 71.33 & ${\uparrow 3.87 }$ & 76.00 & ${\uparrow 8.59 }$ \\ 
& PPR & 82.00 & ${\uparrow 3.36 }$ & 70.67 & ${\uparrow 2.91 }$ & 74.67 & ${\uparrow 6.69 }$ \\ 
\bottomrule[1.5pt]
\end{tabular}
}
\caption{The accuracy of the LLM in solving node classification task across various orders, datasets, and sampling methods. ${\uparrow}$ indicates the improvement compared to the baseline under the same setting.}
\label{table:3}
\end{table*}

\paragraph{(Q2) Is the robustness of LLM to graph description order consistent across different tasks?}
As presented in Figure~\ref{pdf:10}, the variance in LLM performance across different graph description orders reveals a clear pattern: simpler tasks, such as connectivity and cycle detection, consistently exhibit low variance, indicating greater robustness in LLM reasoning. In contrast, more complex tasks like Hamilton path, topological sort, and shortest path show significantly higher variance, reflecting their increased sensitivity to graph description order. Notably, the shortest path task has the highest variance, as it is the only task that requires reasoning on weighted graphs, where changes in graph description greatly affect task complexity.

The robustness in simpler tasks likely stems from their reliance on local substructures, minimizing the need for global reasoning and allowing LLMs to focus on individual graph components without considering the overall structure. As a result, even when the graph description order varies, the model can extract the necessary information with minimal disruption. In contrast, the higher variance in more complex tasks can be attributed to the need for global reasoning, which is heavily influenced by the order of the input sequence. Counterintuitively, even when using CoT to encourage LLMs to engage in rational thinking and provide reasoning steps, the variance did not significantly decrease. A reason for this phenomenon could be that CoT encourages the LLMs for "slow thinking" about the question, but does not necessarily mitigate the attention bias to graph structure during CoT reasoning steps \citep{wang2023primacy}.

\begin{figure*}[t] \vspace{-0.5cm}
    \centering
    \subfloat[Connectivity]{\includegraphics[width=0.32\textwidth]{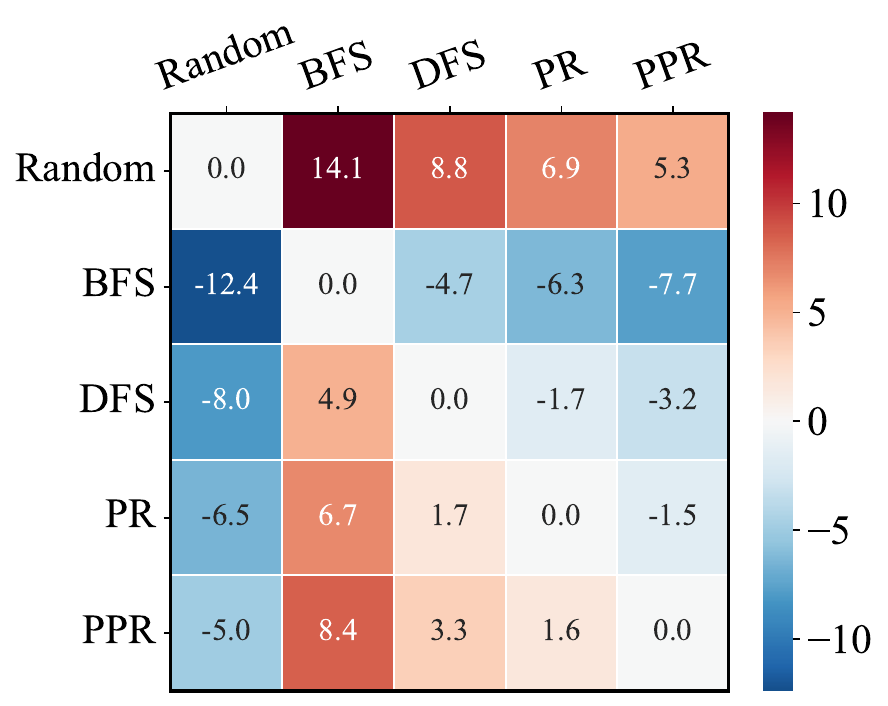}} \hspace{0.005\textwidth}
    \subfloat[Hamilton Path]{\includegraphics[width=0.32\textwidth]{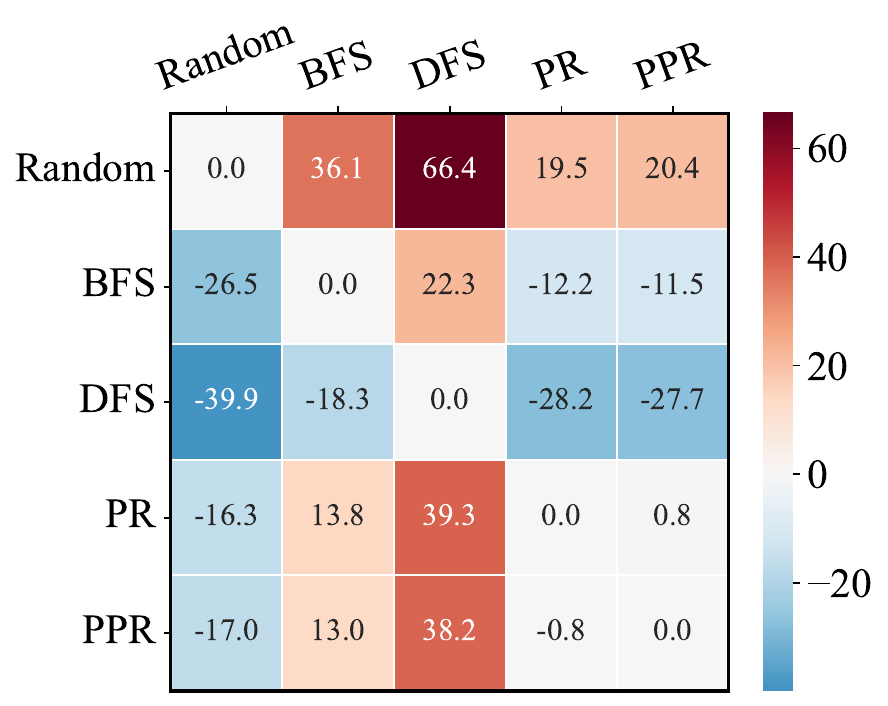}} \hspace{0.005\textwidth}
    \subfloat[Shortest Path]{\includegraphics[width=0.32\textwidth]{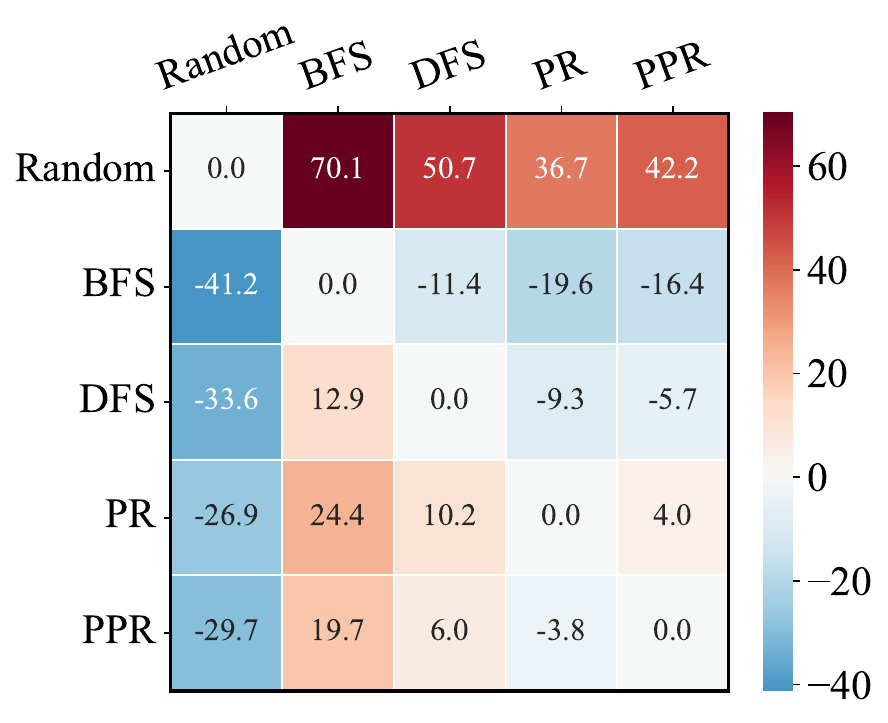}} \\[0.05cm] 
    \caption{The improvement of average accuracy (calculated as the mean across all prompt types) of the LLM between a graph description in one order (horizontal axis) and its average accuracy on graph descriptions in other orders (vertical axis).}
    \label{fig:20-21-22}
\end{figure*}

\begin{figure}[t] 
    \centering
    \includegraphics[width=1.0\linewidth]{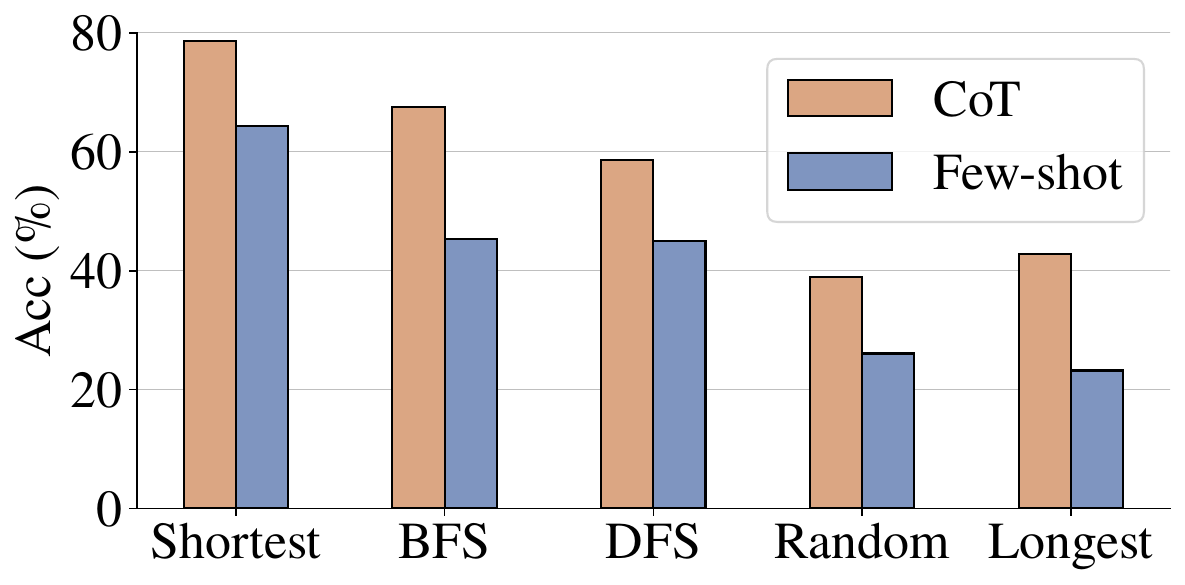} 
    \caption{Results of the accuracy of various orders on shortest path task.} 
    \label{pdf:4}
\end{figure}

\paragraph{(Q3) Does a specific graph description order favor certain graph tasks?}

As presented in Table~\ref{table:2}, BFS generally outperforms DFS in tasks like cycle detection, connectivity detection, and shortest path. For example, BFS achieves 72.71\% accuracy in cycle detection, 7.04\% higher than DFS achieves 67.93\%. Conversely, for tasks requiring deeper exploration, such as topological sort and Hamilton path, DFS performs better.

Figure~\ref{fig:20-21-22} provides further insights. In connectivity, BFS exceeds the random baseline by 14.1\%, while DFS improves upon it by 8.8\%. In the shortest path task, BFS improves accuracy by 70.1\%, outperforming DFS by 12.9\%. Conversely, in the Hamilton path task, DFS surpasses the random order by 66.4\% and outperforms BFS by 22.3\%.

Tasks like cycle and connectivity detection, which focus on local connectivity, benefit from BFS’s level-wise traversal, allowing the LLM to efficiently extract adjacent connections and form accurate local representations. For shortest path problems, BFS ensures the shortest path is found once the target node is reached. In contrast, tasks like Hamilton path and topological sort require a deeper understanding of global structures, where DFS excels by thoroughly exploring paths and capturing global dependencies.

For node classification task, as demonstrated in Table~\ref{table:3}, the PR order consistently outperforms the PPR across all datasets, while PPR generally performs better than traversal-based orders. 

We hypothesize that when LLMs reason based on PR-ordered graph descriptions, their focus on local features may lead to overfitting to the local neighborhood, thereby limiting their ability to capture broader patterns. Additionally, node classification task often require a more comprehensive understanding of the global graph structure, as nodes within the same category may be distributed across multiple regions, making local information inadequate for accurate classification. As for why probability distribution-based orders outperform traversal-based orders, we believe it is because the classification of the query node is influenced by its neighboring nodes, making it less suitable to infer from traversal-based graph descriptions.

\section{Deeper Exploration}
\paragraph{Better graph understanding or just more overlap with the answer?} Inspired by the finding of Q3, we considered whether the improved LLM performance on graph problems is due to the ordered graph descriptions containing all or part of the ground truth. For example, in the shortest path problem, the BFS and DFS edge lists may partially overlap with the correct shortest path.

To validate this hypothesis, We designed two extreme orders for shortest path task:
\begin{itemize} [noitemsep,topsep=0pt,parsep=0pt,partopsep=0pt,leftmargin=1em]
    \item  \textbf{Shortest Path Order}:  Edges are ordered based on the shortest path from the root node \( v_0 \) to the target node \( v_t \). 
    \item \textbf{Longest Path Order}:  Edges are ordered according to the longest path from \( v_0 \) to \( v_t \). 
\end{itemize}

We test the two orders on a subset of GraphDO, and the results are shown in Figure \ref{pdf:4}. The shortest path order, which has the highest overlap with the answer, achieves 78.57\% accuracy with the CoT prompt, a 16.4\% improvement over BFS. In contrast, the accuracy of the longest path order is nearly identical to the random order. Although the shortest path order shows significant improvement, it still falls significantly short of 100\%, indicating that while overlap with the answer has some influence, it is not the sole factor. This confirms that ordered graph descriptions can indeed enhance LLMs' understanding of graphs.

\begin{figure*}[t] \vspace{-0.5cm}
    \centering
    \subfloat[Connectivity]{\includegraphics[width=0.33\linewidth]{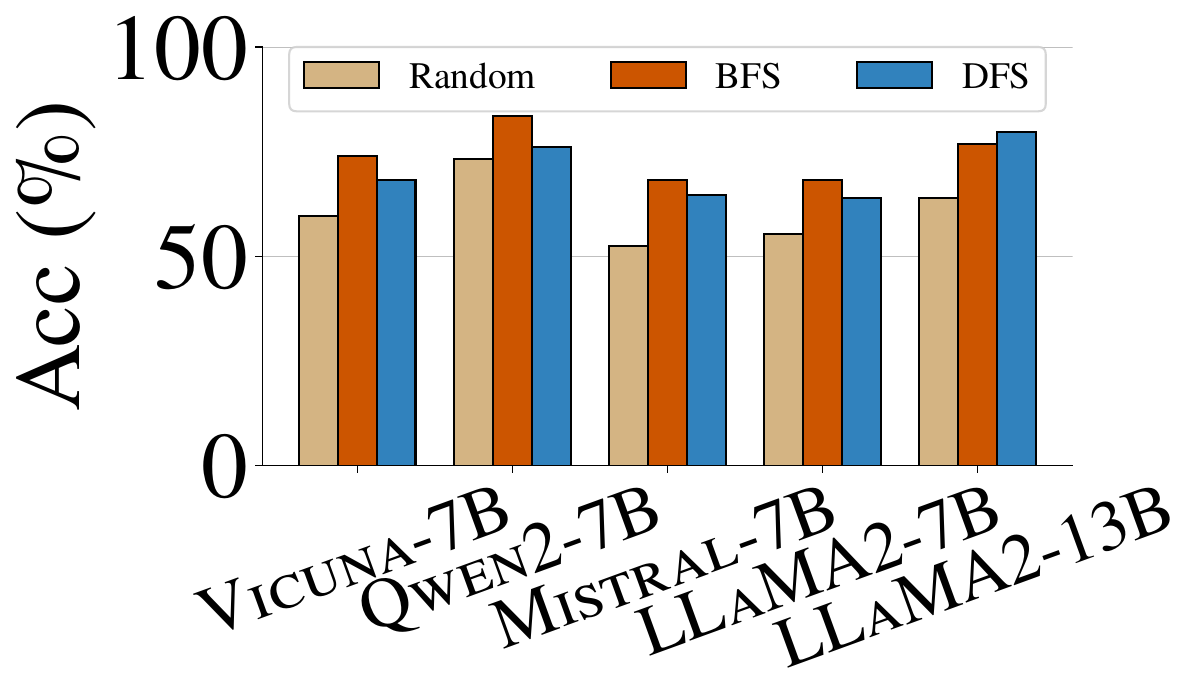}} 
    \subfloat[Cycle]{\includegraphics[width=0.33\linewidth]{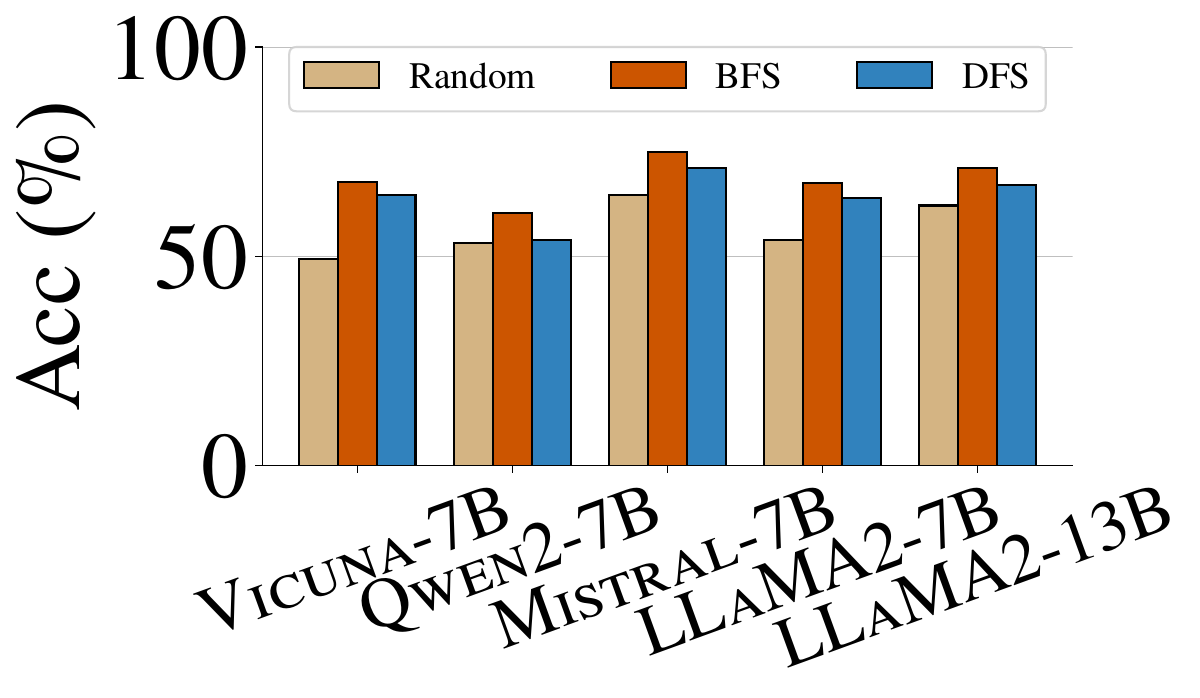}} 
    \subfloat[Shortest Path]{\includegraphics[width=0.33\linewidth]{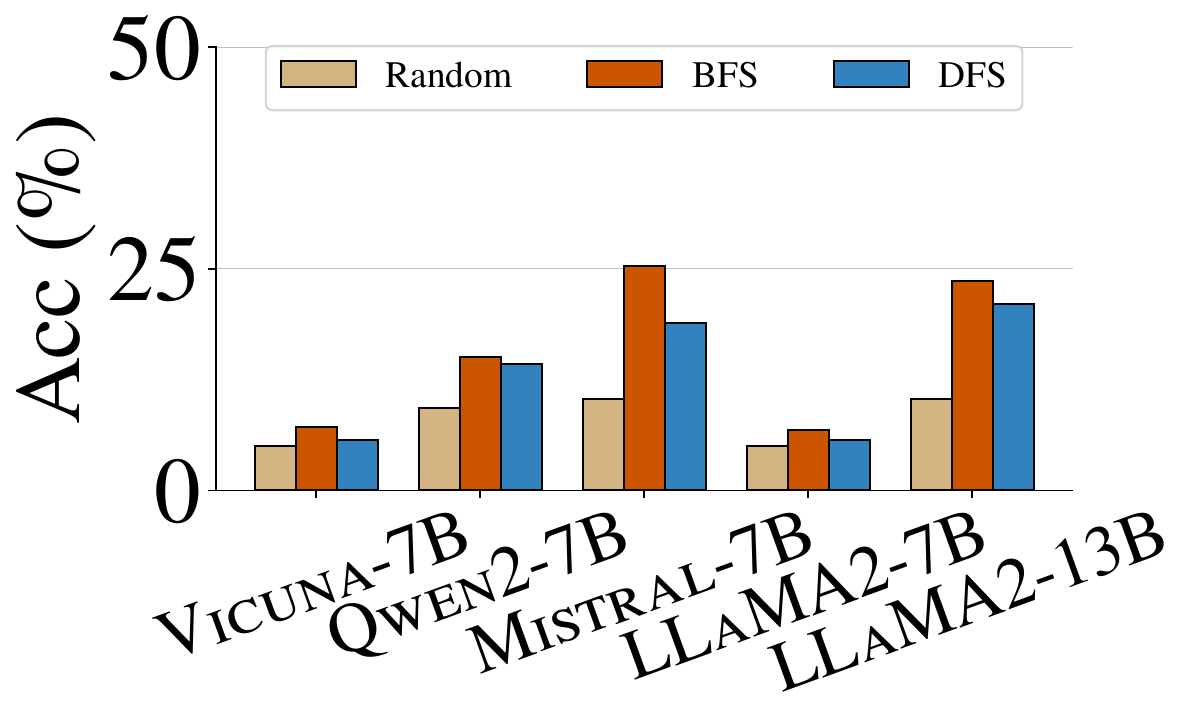}} \\[0.05cm]
    
    \subfloat[Hamilton Path]{\includegraphics[width=0.45\linewidth]{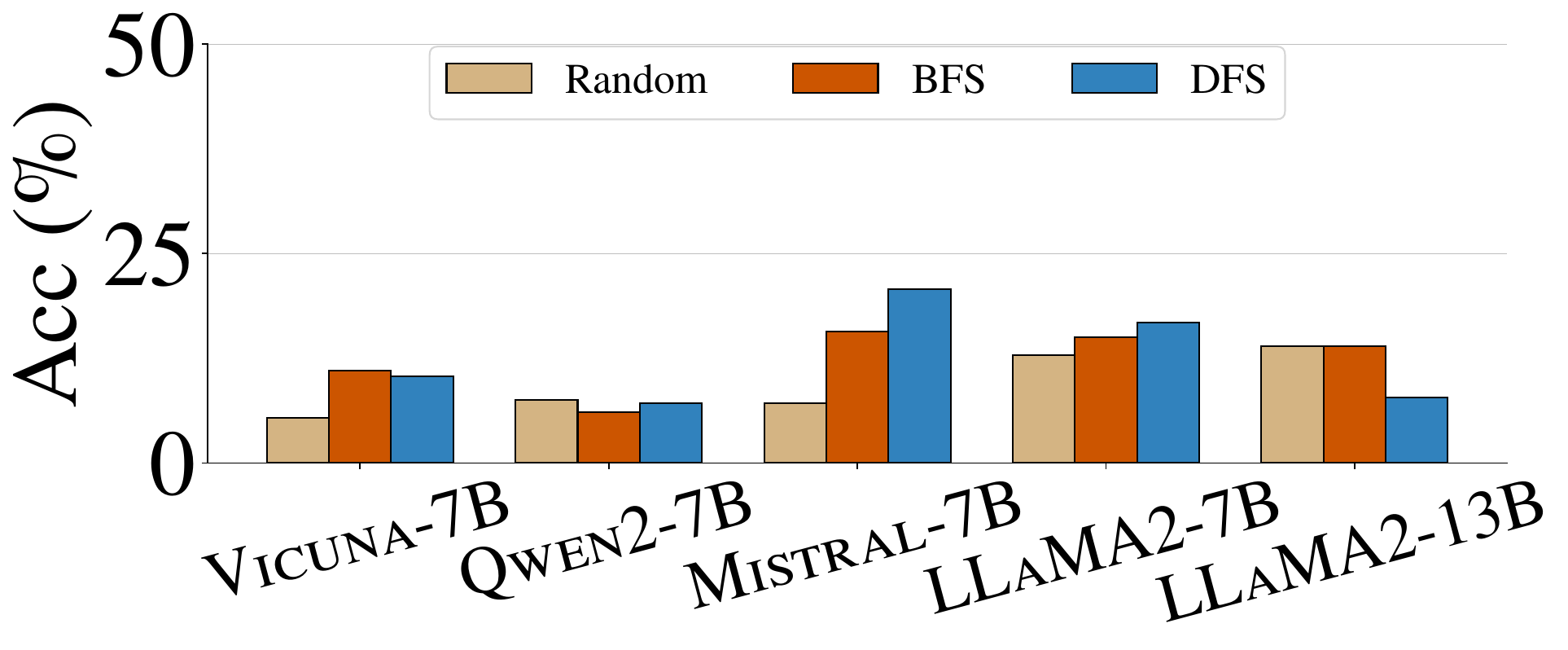}} \hspace{0.03\textwidth}
    \subfloat[Topological Sort]{\includegraphics[width=0.45\linewidth]{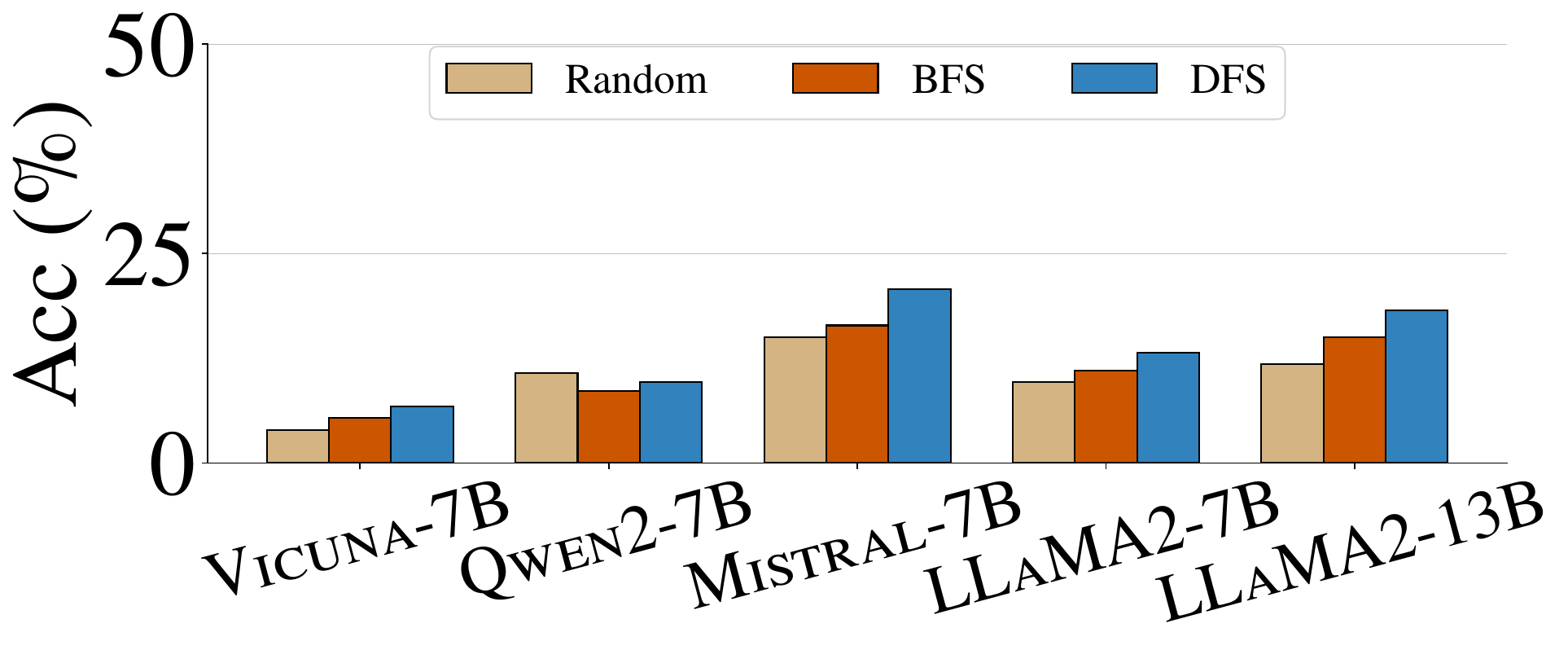}} 
    \caption{The impact of model differences on solving graph reasoning problems.}
    \label{fig:7}
\end{figure*}

\paragraph{Model Comparison Study}
We repeat a subset of the experiments from Table \ref{table:2} on five open-source LLMs to test the generalizability of our findings. As demonstrated in Figure \ref{fig:7}, similar patterns emerged, consistent with previous results, though the effects were less pronounced than with \gpta. We attribute this to the relatively weaker reasoning abilities of these open-source models and their limited capacity to map textual graph descriptions to conceptual spaces \citep{patel2021nlp}. Consequently, graph order has a smaller impact on these models compared to \gpta. Additionally, some models demonstrate superior performance in specific tasks. For example, although \qwen does not excel in other tasks, it shows outstanding performance in the connectivity task, even surpassing \llamab with larger capacity.

        \section{Related Work}
\paragraph{LLMs Reasoning}
LLMs' reasoning and common-sense skills are applied to decision-making and action tasks in various domains. \citet{yao2022react} proposed ReAct, a prompting method that synergizes thinking and action, making it particularly effective for more complex problems involving autonomous planning and exploration. \citet{madaan2023self} introduced Self-Refine, a feedback-driven iterative refinement approach to rectify the mistakes and hallucinations during the inference time of LLM reasoning. \citet{shinn2023reflexion} proposed Reflexion, which uses linguistic feedback stored in episodic memory to enhance decision-making in language agents. \citet{sun2023adaplanner} presented AdaPlanner, a closed-loop approach that enables LLM agents to refine their plans adaptively in response to environmental feedback, integrating in-plan and out-of-plan strategies to improve sequential decision-making. More recently, \citet{zhou2023language} introduced Language Agent Tree Search (LATS), a framework that combines Monte Carlo Tree Search with LM-powered value functions and self-reflections, enabling more deliberate and adaptive decision by integrating reasoning, acting, and planning.

\paragraph{Graph Reasoning with LLMs}
\citet{wang2023can} introduced NLGraph, a benchmark of graph problems in natural language, and proposed Build-a-Graph and Algorithmic prompting to improve LLM performance. \citet{fatemi2023talk} conducted the first comprehensive study on encoding graph-structured data as text for LLMs, revealing that task performance depends on encoding methods, graph tasks, and graph structure. \citet{zhao2023graphtextgraphreasoningtext} presented graphtext, a framework that converts graphs into natural language using a graph-syntax tree, enabling facilitating interactive communication between humans and LLMs. \citet{wei2024gitagraphvisualtextual} proposed GITA, an end-to-end framework that integrates visual graphs into general graph reasoning. \citet{das2023modality} encode a graph with diverse modalities to enhance LLM efficiency in processing complex graph structures. 

\section{Conclusion}
In this work, we conduct the first comprehensive analysis of how graph description order affects LLM performance in solving graph problems. Our findings demonstrate that ordered graph descriptions significantly enhance LLMs' ability to comprehend and reason about graph structures, a crucial discovery that could reshape the way we approach graph reasoning tasks. Additionally, through detailed analysis of various graph description orders, we observe that the impact of order on performance is closely tied to the intrinsic characteristics of each task. We believe that the over-reliance on graph descriptions stems from limitations in positional encoding—what we refer to as attention bias. Lastly, we introduce the GraphDO dataset, which aims to advance the community’s understanding of how graph descriptions influence reasoning in LLMs, providing a valuable benchmark for future research in this area.

\clearpage
\section*{Limitations}
While our work demonstrates the critical role that the order of graph descriptions plays in LLMs' understanding of graphs, we have not explored the impact of this order on different graph structures and types in greater depth. Additionally, although we conducted some in-depth analyses, we did not provide a rigorous mathematical and theoretical explanation for the phenomena observed in this paper, which warrants further experimental investigation.

\section*{Ethics Statement}
In conducting our research, we place paramount importance on ethical standards to ensure integrity and contribute positively to the scientific community. We exclusively utilize open-source datasets, ensuring that our work is built upon accessible and transparent resources. Our methods employ models that are either open-source or have gained wide recognition for their reliability and ethical use within the academic community. Furthermore, we have meticulously designed our methodology to prevent the generation of harmful or misleading information, thereby safeguarding the integrity of our findings.

\section*{Acknowledgements}
This work is supported in part by the National Key R\&D Program of China under Grant Nos. 2023YFA1011602 and 2023YFC3305303, and the National Natural Science Foundation of China under Grant Nos. 62472408, 62372431, 62441229, and 62377043.

        \nocite{*}
        \bibliography{reference}
        \clearpage
\appendix

\begin{table*}[h]
    \centering
    \small
    \renewcommand{\arraystretch}{0.8}
    \begin{tabularx}{\textwidth}{l X} 
        \toprule
        \textbf{Prompt Style} & \textbf{Prompt Template} \\
        \midrule
        Zero-shot & Graph: \textcolor{black}{<Graph Description>} \textcolor{gray}{\textbackslash n} Question: \textcolor{black}{<Question>} \textcolor{gray}{\textbackslash n} Answer: \\
        \midrule
        Zero-shot CoT & Graph: \textcolor{black}{<Graph Description>} \textcolor{gray}{\textbackslash n} Question: \textcolor{black}{<Question>} Let's think step by step. \textcolor{gray}{\textbackslash n} Answer: \\
        \midrule
        Few-shot & Graph: \textcolor{black}{<Example Graph Description>} \textcolor{gray}{\textbackslash n} Question: \textcolor{black}{<Example question>} Answer: \textcolor{black}{<Example Answer>} \textcolor{gray}{\textbackslash n} ... (more few-shot examples) \textcolor{gray}{\textbackslash n} Graph: \textcolor{black}{<Graph Description>} \textcolor{gray}{\textbackslash n} Question: \textcolor{black}{<Question>} \textcolor{gray}{\textbackslash n} Answer: \\
        \midrule
        CoT & Graph:\ textcolor{black}{<Example Graph Description>} \textcolor{gray}{\textbackslash n} Question: \textcolor{black}{<Example Question>} Answer: \textcolor{black}{<Example Answer with CoT>} \textcolor{gray}{\textbackslash n} ... (more examples with CoT) \textcolor{gray}{\textbackslash n} Graph: \textcolor{black}{<Graph Description>} \textcolor{gray}{\textbackslash n} Question: \textcolor{black}{<Question>} \textcolor{gray}{\textbackslash n} Answer: \\
        \midrule
        CoT-BAG & Graph: \textcolor{black}{<Example Graph Description>} \textcolor{gray}{\textbackslash n} Question: \textcolor{black}{<Example Question>} Answer: \textcolor{black}{<Answer with CoT>} \textcolor{gray}{\textbackslash n} ... (more examples with CoT) \textcolor{gray}{\textbackslash n} Graph: \textcolor{black}{<Graph Description>} \textcolor{gray}{\textbackslash n} Question: \textcolor{black}{<Question>} \textcolor{gray}{\textbackslash n} Let's construct a graph with the nodes and edges first \textcolor{gray}{\textbackslash n} Answer: \\

        \bottomrule
    \end{tabularx}
    \caption{Prompt styles and their corresponding templates for graph reasoning tasks.}
    \label{table:1}
\end{table*}

\section{Additional experimental details}

\subsection{Prompt Design} \label{sec:prompt}
We employe various prompting methods in our experiments. We briefly introduce these methods here:

\begin{itemize} [noitemsep,topsep=0pt,parsep=0pt,partopsep=0pt,leftmargin=1em]
    \item \textbf{Zero-shot}: Zero-shot prompting only provides the task description and uses zero exemplars, requiring the model to generate the desired output.
    \item \textbf{Zero-shot CoT} \citep{wei2022chain}: Zero-shot CoT prompting involves appending a thought inducing phrase "Let’s think step by step." 
    \item \textbf{Few-shot}\citep{brown2020language} : Few-shot prompting provides the LLM with a few exemplars, including task descriptions and expected outputs, to guide its learning.
    \item \textbf{Chain-of-Thought (CoT)} \citep{wei2022chain}: CoT prompting provides the model with a series of exemplars, each demonstrating a step-by-step approach to solving the task. It encourages the LLM to articulate its reasoning process before presenting the final answer.
    \item \textbf{CoT-BAG}\citep{wang2023can}: Similar to CoT, but CoT-BAG prompting appends the phrase "Let’s construct a graph with the nodes and edges first" at the end of the text to guide the model's reasoning process.
\end{itemize}

A more detailed explanation can be found in Table \ref{table:1}.

\subsection{Personalization Vector} \label{pv}

For the Personalized PageRank (PPR) order, the personalization vector \( e_v \) is defined in a task-specific manner. The definitions for various tasks are as follows:

\paragraph{T1 Connectivity} For the two queried nodes \( u \) and \( v \), the personalization vector is defined as:
\[
e_u = e_v = 0.5 \quad \text{and} \quad e_{w} = 0 \quad \forall w \in \mathcal{V} \setminus \{u, v\}.
\]

\paragraph{T2 Cycle} If a cycle \( \mathcal{C} \) exists, the personalization vector \( e_v \) is uniformly distributed across the nodes forming the cycle, with the sum equal to 1:
\[
e_v = \frac{1}{|\mathcal{C}|} \quad \forall v \in \mathcal{C}, \quad e_w = 0 \quad \forall w \in \mathcal{V} \setminus \mathcal{C}.
\]
If no cycle exists, \( e_v \) is uniformly distributed across all nodes:
\[
e_v = \frac{1}{|\mathcal{V}|} \quad \forall v \in \mathcal{V}.
\]

\paragraph{T3 Hamiltonian Path} For the nodes along the Hamiltonian path \( \mathcal{H} \), the personalization vector is uniformly distributed, with the sum equal to 1:
\[
e_v = \frac{1}{|\mathcal{H}|} \quad \forall v \in \mathcal{H}, \quad e_w = 0 \quad \forall w \in \mathcal{V} \setminus \mathcal{H}.
\]
Here, \( \mathcal{H} \) represents the set of nodes on the Hamiltonian path.

\paragraph{T4 Shortest Path} For the nodes along the shortest path \( \mathcal{P}_{uv} \) between two nodes \( u \) and \( v \), the personalization vector is uniformly distributed, with the sum equal to 1:
\[
e_v = \frac{1}{|\mathcal{P}_{uv}|} \quad \forall v \in \mathcal{P}_{uv}, \quad e_w = 0 \quad \forall w \in \mathcal{V} \setminus \mathcal{P}_{uv}.
\]
Here, \( \mathcal{P}_{uv} \) represents the set of nodes on the shortest path between nodes \( u \) and \( v \).

\paragraph{T5 Topological Sort} For the nodes with in-degree 0 in a directed acyclic graph, the personalization vector is uniformly distributed, with the sum equal to 1:
\[
e_v = \frac{1}{|\mathcal{V}_0|} \quad \forall v \in \mathcal{V}_0, \quad e_w = 0 \quad \forall w \in \mathcal{V} \setminus \mathcal{V}_0,
\]
where \( \mathcal{V}_0 \) represents the set of nodes with in-degree 0.

\paragraph{T6 Node Classification} The personalization vector is defined based on the shortest path distance \( \delta(v) \) from the target node \( v_0 \) to each node \( v \), with the formula:
\[
e_v = \frac{\Delta - \delta(v) + 1}{\sum_{u \in \mathcal{V}} (\Delta - \delta(u) + 1)},
\]
where \( \delta(v) \) is the shortest path distance from node \( v_0 \) to node \( v \), and \( \Delta \) represents the maximum shortest path distance from \( v_0 \) to any node in the graph.

\subsection{Response Parser} \label{sec:parser}
For GPT series models, we utilize string matching to parse responses. In binary classification tasks, such as cycle detection and connectivity detection, we extract answers by matching specific keywords like "there is a cycle" or "there is no cycle" to assess correctness. Path-related tasks are more complex: we first locate the approximate position of the path by matching keywords such as "the shortest path from x to x," then use numerical matching to extract the node indices and evaluate whether the path satisfies the task requirements.

However, for open-source LLMs, the weaker in-context learning abilities of smaller models compared to GPT make it difficult for them to replicate the provided example responses, complicating answer extraction through string matching. Additionally, differences in training data and methods often result in distinct response styles across LLMs, further hindering the use of parsers. Designing a custom parser for each LLM would significantly increase the workload. Therefore, we employe GPT to verify whether the responses of open-source LLMs align with the correct answers.

\section{GraphDO} \label{sec:odgraph}
\subsection{Graph Generation}
For traditional graph tasks, we employe the Erdős-Rényi (ER) graph generation method. Specifically, we set the number of nodes \(n\) and a connection probability \(p\), where any two nodes are connected with probability \(p\). Edges can be directed or undirected based on the task. In our experiments, we chose \(n\) between 5 and 15, as previous studies have shown that LLMs demonstrate more consistent reasoning abilities on graphs of this size, making them suitable for detecting patterns in LLM performance \citep{wang2023can, cao2024graphinsightunlockinginsightslarge}. The connection probability \(p\) was fixed at 0.3 to ensure that the graphs had a moderate level of sparsity, which is crucial for evaluating the reasoning capabilities of LLMs without generating overly dense or trivial graph structures.
Additionally, given the diversity of the tasks, we filtered the generated graphs to guarantee that each instance had a valid and well-defined solution. For example, in the shortest path problem, we ensure that a valid path always exists between the start and end nodes.

\subsection{Sampling}
For node classification task, given that the data is sourced from real-world datasets, graph sampling is necessary. Below, we outline the two sampling methods used.

\paragraph{Ego} Starting from a selected central node, we expand the subgraph by including neighboring nodes within a fixed hop distance of 3, ensuring that the resulting subgraph contains 50 nodes.

\paragraph{Forest Fire} Propagating from a selected seed node to neighboring nodes with a probability of 0.3 continues until 50 nodes are included or no further propagation is possible.

For the sampled subgraphs, the graphs are converted into text-based descriptions using various orders.

\subsection{Statistics}
Using the above methodology, we collecte a total of 1,700 graphs ($280 \times 5 + 50 \times 3 \times 2$). We encode the graphs in adjacency format and then sorted the edges according to different orders, converting them into text sequences in natural language. This resulted in 8,500 graph cases, with each case consisting of a graph description, a question, and an answer.

\section{Implementation Details}
We deploy the open-source LLMs for our experiments on a 4 × A800 server. The decoding temperature was set to zero. We use the NetworkX library \citep{hagberg2008exploring} to generate the random graphs and to find the answers to the graph problems. The node classification experiments presented in Table \ref{table:3} were repeated three times due to the limited sample size and the prompt is set as zero-shot. When calculating PageRank score, the default weight of edges in an unweighted graph is set to 1.

\section{Case Analysis} \label{sec:case}

As shown in Table \ref{table:4} and Table \ref{table:5}, although GPT answered most of the questions correctly, its imitation of the CoT provided in the prompt was incomplete, demonstrating limited in-context learning ability. In the erroneous examples presented in Table \ref{table:5}, GPT exhibited reasoning that appeared logically sound, but factual errors were made, which can be attributed to the phenomenon of hallucination in LLMs.

\begin{table*}[!h] 
    \centering
    \small
    \setlength{\tabcolsep}{10pt}
    \setlength{\extrarowheight}{4pt} 
    \resizebox{\textwidth}{!}{
    \begin{tabular}{p{4in}|p{2in}}
     \toprule[1.5pt]
     {\textbf{Question}} & \textbf{Answer of gpt-3.5-0613}  \\
     \midrule[1pt]
    Graph: In an undirected graph, (i, j) means that node i and node j are connected with an edge, and the edges are: (0,1) (0,2) (1,5) (1,2) (1,3) (2,5). \newline
    Question: Determine if there is a path between two nodes in the graph. Is there a path between node 2 and node 3? \newline
    Answer: Node 2 is connected to node 1, node 1 is connected to node 3. We can follow the path: 2->1->3, so the answer is yes. \newline
    
    Graph: In an undirected graph, (i, j) means that node i and node j are connected with an edge, and the edges are: (0,2) (0,5) (0,6) (1,3) (1,4) (2,5) (2,6) (3,4) (5,6). \newline
    Question: Determine if there is a path between two nodes in the graph. Is there a path between node 5 and node 4? \newline
    Answer: Node 5 is in the connected block consisted of node 0, node 2, node 5, and node 6. Node 4 is in the connected block consisting of node 1, node 3, and node 4. Node 5 and node 4 are not in the same connected block, so the answer is no. \newline

    Graph: In an undirected graph, (i, j) means that node i and node j are connected with an edge, and the edges are: (4,0) (4,1) (4,2). \newline
    Question: Determine if there is a path between two nodes in the graph. Is there a path between node 0 and node 2? \newline
    Answer: 

    & Node 0 is connected to node 4, and node 4 is connected to node 2. We can follow the path: 0->4->2, so the answer is yes. \CheckmarkBold \\
    
    \cline{1-2}

    Graph: In an undirected graph, (i,j) means that node i and node j are connected with an edge, and the edges are: (3,4) (3,5) (1,0) (2,5) (2,0). \newline Question: Is there a cycle in this graph? \newline Answer: No, there is no cycle in this graph. \newline 
    
    Graph: In an undirected graph, (i,j) means that node i and node j are connected with an edge, and the edges are: (3,5) (1,0) (3,0) (3,4) (4,1) (2,3). \newline Question: Is there a cycle in this graph? \newline Answer: The edges (3,0), (1,0), (4,1), (3,4) form a cycle, so yes, there is a cycle in this graph. \newline 
    
    
    Graph: In an undirected graph, (i,j) means that node i and node j are connected with an edge, and the edges are: (1,3) (1,0) (4,0) (4,2). \newline Question: Is there a cycle in this graph? \newline 
    Answer:

    & No, there is no cycle in this graph. The graph forms a tree-like structure without any closed loops. \CheckmarkBold \\
     \cline{1-2}

     Graph: In an undirected graph, (i, j, w) means that nodes i and j are connected by an edge with weight w, and the edges are: (0, 1, 2) (1, 4, 1) (0, 2, 1) (2, 3, 1) (3, 4, 2). \newline Question: Give the shortest path from node 0 to node 4. \newline Answer: All the paths from node 0 to node 4 are: 0,1,4 with a total weight of 2 + 1 = 3, 
     
     0,2,3,4 with a total weight of 1 + 1 + 2 = 4. 
     
     The weight of path 0,1,4 is the smallest, so the shortest path from node 0 to node 4 is 0,1,4 with a total weight of 3. \newline 
     
     Graph: In an undirected graph, (i, j, w) means that nodes i and j are connected by an edge with weight w, and the edges are: (0, 3, 2) (0, 4, 1) (0, 2, 1) (4, 1, 2) (2, 1, 1) (3, 2, 4) (2, 4, 1) (3, 4, 2). \newline Question: Give the shortest path from node 3 to node 1. \newline Answer: All the paths from node 3 to node 1 are: 3,2,1 with a total weight of 4 + 1 = 5, 
     
     3,2,4,1 with a total weight of 4 + 1 + 2 = 7, 
     
     3,4,1 with a total weight of 2 + 2 = 4, 
     
     3,4,2,1 with a total weight of 2 + 1 + 1 = 4, 3,0,4,1 with a total weight of 2 + 1 + 2 = 5, 
     
     3,0,2,1 with a total weight of 2 + 1 + 1 = 4, 3,4,2,4,1 with a total weight of 2 + 1 + 1 + 2 = 6. 
     
     The weight of path 3,4,1 is the smallest, so the shortest path from node 3 to node 1 is 3,4,1 with a total weight of 4. \newline 
     
     Graph: In an undirected graph, (i, j, w) means that nodes i and j are connected by an edge with weight w, and the edges are: (0, 4, 4) (0, 3, 3) (0, 1, 3) (1, 4, 4) (2, 4, 2) (2, 3, 1) (3, 4, 1). \newline Question: Give the shortest path from node 0 to node 2.

    & To find the shortest path from node 0 to node 2 in the given graph, we can use Dijkstra's algorithm. Let's calculate the shortest path step by step: \newline
    
    * Start at node 0.
     
    * From node 0, we can go to node 1 with weight 3 and to node 3 with weight 3.
     
    * From node 1, we can go to node 4 with weight 4.
     
    * From node 3, we can go to node 2 with weight 1. \newline

     Therefore, the shortest path from node 0 to node 2 is 0 → 3 → 2 with a total weight of 3 + 1 = 4.
    \CheckmarkBold \\

    \bottomrule[1.5pt]
    \end{tabular}
    }
    \caption{Cases of the questions and the responses of \gpta.}
    \label{table:4}
\end{table*}

\begin{table*}[!h] 
    \centering
    \small
    \setlength{\tabcolsep}{10pt}
    \setlength{\extrarowheight}{4pt} 
    \resizebox{\textwidth}{!}{
    \begin{tabular}{p{4in}|p{2in}}
     \toprule[1.5pt]
     {\textbf{Question}} & \textbf{Answer of gpt-3.5-0613} \\
     \midrule[1pt]
    Graph: In a directed graph, (i, j) means that node i and node j are connected with an edge, and the edges are: (0, 4), (0, 1), (2, 1), (3, 2). \newline
    Question: Give any topological sorting of the graph. \newline
    Answer: First, I need to find the nodes with an in-degree of 0: 0,3. Then, I will traverse these nodes in sequence: starting from node 0, I traverse nodes 4, 1. Since node 4 and node 1 depend on node 0, after node 0 is visited, nodes 4 and 1 can be visited. Starting from node 3, I traverse node 2. Since node 2 depends on node 3, after node 3 is visited, node 2 can be visited. So the topological sequence of the graph is 0, 3, 4, 2, 1.\newline
    \newline
    Graph: In a directed graph, (i, j) means that node i and node j are connected with an edge, and the edges are: (0, 1), (1, 2), (3, 1), (3, 2), (4, 3), (4, 0), (4, 1). \newline
    Question: Give any topological sorting of the graph. \newline
    Answer: First, I need to find the nodes with an in-degree of 0: 4. Then, I will traverse these nodes in sequence: starting from node 4, I traverse nodes 0, 1, 3. Since node 0 and node 3 both depend on node 4, after node 4 is visited, nodes 0 and 3 can be visited. Since visiting node 1 requires visiting nodes 0 and 3 first, after node 0 and node 3 are visited, node 1 can be visited. Since visiting node 2 requires visiting nodes 1 and 3 first, after node 1 and node 3 are visited, node 2 can be visited. So the topological sequence of the graph is 4, 0, 3, 1, 2. \newline

    Graph: In a directed graph, (i, j) means that node i and node j are connected with an edge, and the edges are: (0, 3), (1, 2), (3, 2), (4, 1), (4, 2), (5, 1), (5, 2).\newline
    Question: Give any topological sorting of the graph.\newline
    Answer: 

    & 
    To find a topological sorting of the given directed graph, we can follow the steps of Kahn's algorithm. Here's how we can do this for the provided graph:
    \newline
    
    * Find nodes with in-degree 0.
    
    * Start with these nodes and remove their outgoing edges.
    
    * Repeat until all nodes are visited.
    
    * For the given graph with edges: (0, 3), (1, 2), (3, 2), (4, 1), (4, 2), (5, 1), (5, 2), the topological sorting could be as follows: 0, 4, 5, 1, 3, 2
    \newline
    
    This sequence satisfies the topological order where each node appears after its dependencies. \CheckmarkBold
    \\
    \cline{1-2}

    Graph: In an undirected graph, (i, j) means that node i and node j are connected with an edge, and the edges are: (4, 2), (0, 4), (4, 3), (0, 1), (0, 2), (4, 1), (2, 3). \newline Question: Is there a path in this graph that visits every node exactly once? If yes, give the path. Note that in a path, adjacent nodes must be connected with edges. \newline Answer: Yes. We can start at node 0. As node 0 is connected with node 1, and node 1 is not visited, we can then visit node 1. As node 1 is connected with node 4, and node 4 is not visited, we can then visit node 4. As node 4 is connected with node 2, and node 2 is not visited, we can then visit node 2. As node 2 is connected with node 3, and node 3 is not visited, we can then visit node 3. Therefore, the path can be: 0, 1, 4, 2, 3. \newline

    Graph: In an undirected graph, (i, j) means that node i and node j are connected with an edge, and the edges are: (0, 4), (1, 5), (3, 1), (4, 2), (3, 5), (2, 1), (1, 4), (2, 5). \newline Question: Is there a path in this graph that visits every node exactly once? If yes, give the path. Note that in a path, adjacent nodes must be connected with edges. \newline Answer: Yes. We can start at node 1. As node 1 is connected with node 3, and node 3 is not visited, we can then visit node 3. As node 3 is connected with node 5, and node 5 is not visited, we can then visit node 5. As node 5 is connected with node 2, and node 2 is not visited, we can then visit node 2. As node 2 is connected with node 4, and node 4 is not visited, we can then visit node 4. As node 4 is connected with node 0, and node 0 is not visited, we can then visit node 0. The path can be: 1, 3, 5, 2, 4, 0. \newline 
    
    Graph: In an undirected graph, (i, j) means that node i and node j are connected with an edge, and the edges are: (0, 2), (0, 4), (0, 1), (1, 3), (1, 2), (3, 4). \newline Question: Is there a path in this graph that visits every node exactly once? If yes, give the path. Note that in a path, adjacent nodes must be connected with edges. \newline Answer:

    &

    Yes, there is a path that visits every node exactly once in this graph.\newline

    We can start at node 0. Then we can visit node 1, as it is connected to node 0. Next, we move to node 3 since it is connected to node 1. After that, we visit node 4 which is connected to node 3. Finally, we can move to node 2 from node 4, completing the path.\newline
    
    Therefore, the path that visits every node exactly once is: 0, 1, 3, 4, 2. \XSolidBrush
    \\

    \bottomrule[1.5pt]
    \end{tabular}
    }
    \caption{Cases of the questions and the responses of \gpta.}
    \label{table:5}
\end{table*}
	
\end{document}